\documentclass{isprs}
\usepackage{subfig}
\usepackage{setspace}
\usepackage{geometry}
\usepackage{epstopdf}
\usepackage[british]{babel} 
\usepackage{color}
\usepackage{cite}
\usepackage{amsmath}
\usepackage{amssymb}

\usepackage{makecell}

\usepackage{blindtext}

\usepackage[pdftex, english, pdfpagelabels=true]{hyperref}
\hypersetup{
    colorlinks = true
}

\begin{document}

\title{
Uncertainty Estimation for End-To-End Learned Dense Stereo Matching via Probabilistic Deep Learning
}

\author{
    Max Mehltretter
}

\address{
	Institute of Photogrammetry and GeoInformation, Leibniz University Hannover, Germany\\
	mehltretter@ipi.uni-hannover.de
}

\commission{}{} 
\workinggroup{} 
\icwg{}   

\abstract{
    Motivated by the need to identify erroneous disparity assignments, various approaches for uncertainty and confidence estimation of dense stereo matching have been presented in recent years. As in many other fields, especially deep learning based methods have shown convincing results. However, most of these methods only model the uncertainty contained in the data, while ignoring the uncertainty of the employed dense stereo matching procedure. 
    Additionally modelling the latter, however, is particularly beneficial if the domain of the training data varies from that of the data to be processed.
    For this purpose, in the present work the idea of probabilistic deep learning is applied to the task of dense stereo matching for the first time. Based on the well-known and commonly employed GC-Net architecture, a novel probabilistic neural network is presented, for the task of joint depth and uncertainty estimation from epipolar rectified stereo image pairs.
    Instead of learning the network parameters directly, the proposed probabilistic neural network learns a probability distribution from which parameters are sampled for every prediction. The variations between multiple such predictions on the same image pair allow to approximate the model uncertainty.
    The quality of the estimated depth and uncertainty information is assessed in an extensive evaluation on three different datasets.
}


\maketitle

\section{Introduction}\label{sec:introduction}

Nowadays, deep learning based algorithms are frequently employed to process high dimensional data in order to accomplish a wide range of complex tasks - often with convincing results. 
As a consequence, initially the precision of these results is rarely questioned and the associated uncertainty is equally rarely determined.
However, it is obviously crucial to be able to assess how trustworthy a result is. This is particularly true for safety-critical applications. Assigning a high level of uncertainty to erroneous predictions can prevent a system from taking wrong decisions with potentially fatal consequences.

Consequently, the estimation of uncertainty is the topic of many recent investigations in the field of computer vision and photogrammetry, especially in the context of deep learning.
This also applies to the task of dense stereo matching, in which depth is determined for every or at least a large majority of pixels within a stereo image pair. 
In principle, depth reconstruction from stereo images can be interpreted as inverse operation to a perspective projection. Since the projection of a 3D scene to a 2D image plane results in a dimensionality reduction, the inverse operation does not have a unique solution in general, characterising it as ill-posed. To determine a solution nevertheless, the identification of point correspondences within at least two images of a stereo pair is a prerequisite in general.
However, especially under challenging conditions, depth reconstruction approaches might not be able to identify the correct correspondences for all pixels. A measure of uncertainty is therefore important in order to assess the quality of the reconstructed depth information.

\begin{figure}[t]
\centering
\subfloat[Reference image]
{
    \includegraphics[width=0.45\linewidth]{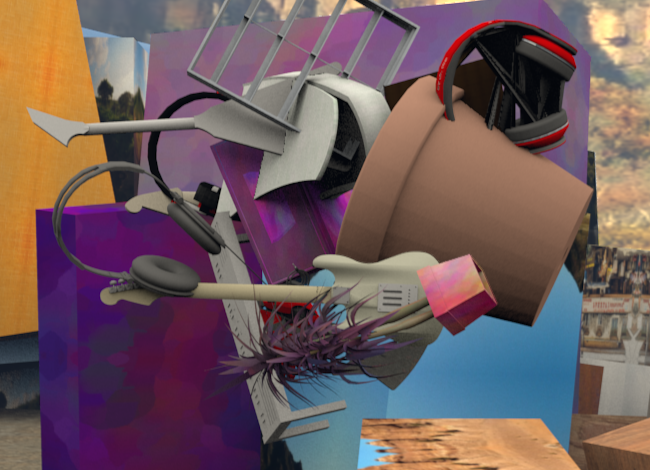}
}\hfill
\subfloat[Error map]
{
    \includegraphics[width=0.45\linewidth]{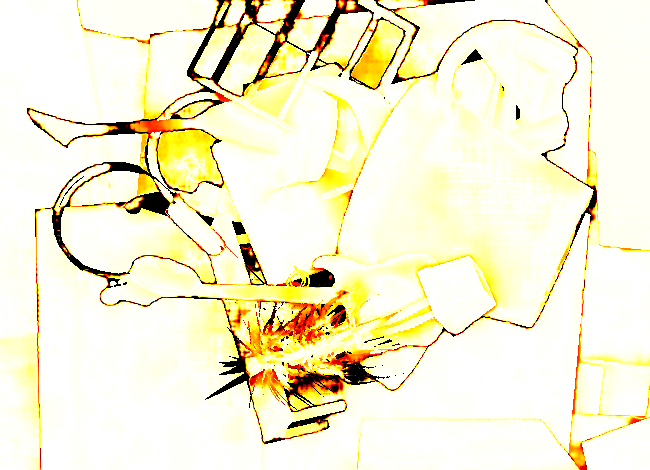}
}\newline
\subfloat[Aleatoric uncertainty]
{
    \includegraphics[width=0.45\linewidth]{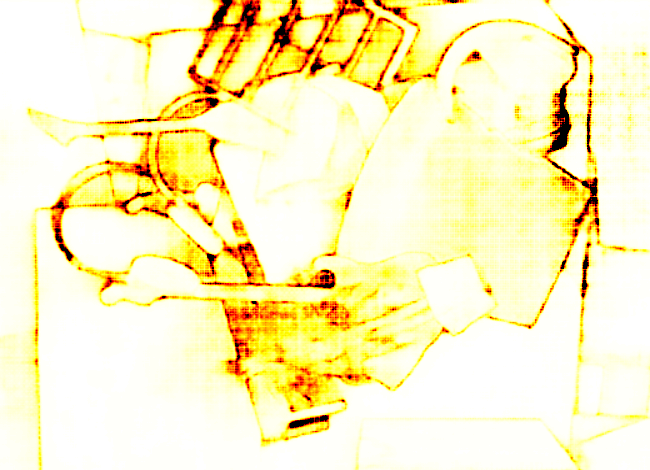}
}\hfill
\subfloat[Combined uncertainty]
{
    \includegraphics[width=0.45\linewidth]{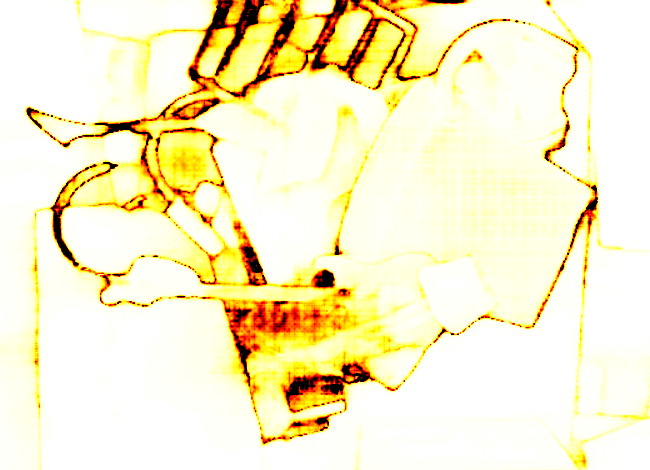}
}
\caption{\textbf{Advantages of modelling epistemic uncertainty.} Compared to aleatoric uncertainty alone, the combination of aleatoric and epistemic uncertainty shows a significantly better correspondence with the real error distribution.}
\label{fig:frontPage}
\end{figure}

However, the majority of approaches for modelling the uncertainty of dense stereo matching only take into account the uncertainty contained in the data, called aleatoric uncertainty. The uncertainty contained in a learned model, called epistemic uncertainty, is thereby neglected. While aleatoric uncertainty indicates image regions which are in general hard to match, such as repetitive patterns or texture-less surfaces, it does not react on samples which are hard to match, since they are too different to the data used for training the network. Epistemic uncertainty on the other hand, is able to indicate patches which are outside of the learned distribution, which is especially valuable if the training and the test domain are different or if the training data does not cover all possible variations of the test domain. The latter is, for example, the case for real-world applications where the test domain is too complex to be covered completely within the training data. 

In order to model both, aleatoric and epistemic uncertainty, in the present work a novel probabilistic neural network is presented.
While depth and aleatoric uncertainty are directly predicted by the network, for the purpose of epistemic uncertainty estimation, the proposed probabilistic neural network learns a probability distribution from which parameters are sampled for every prediction, instead of learning the network parameters directly. Based on the variations between multiple such predictions, the epistemic uncertainty is approximated.
The evaluation of the proposed network is two-folded: On the one hand, the quality of the estimated uncertainty is assessed. On the other hand, the impact on the accuracy of the depth predictions is evaluated and compared against a deterministic baseline.
Thus, the main contributions of this work are:
\begin{itemize}
	\item A neural network architecture allowing to jointly estimate dense depth and uncertainty from epipolar rectified stereo image pairs, based on probabilistic deep learning.
	\item An investigation of the effects of probabilistic deep learning on an end-to-end learned network trained for the task of dense stereo matching.
	\item An extensive evaluation of the behaviour of the modelled uncertainty on the training domain as well as on two additional well established datasets.
\end{itemize}
\section{Related Work}\label{sec:relatedWork}

Within this section, first current approaches for obtaining depth from a stereo image pair are reviewed. This serves as a foundation for the justification for the choice of the network architecture used later. Subsequently, the state of the art in uncertainty estimation in the context of dense stereo matching is discussed.

\subsection{Depth Estimation}

Motivated by the superior accuracy compared to classical approaches, such as Semi-Global Matching \cite{Hirschmuller2008} or ELAS \cite{Geiger2010}, deep learning based methods, carrying out the task of dense stereo matching, gained popularity in recent years.
While some of those methods map single parts of the dense stereo matching pipeline \cite{Scharstein2002} to a neural network, such as matching cost computation \cite{Zbontar2016} and disparity optimisation \cite{Seki2017}, others learn the whole process end to end.

In \cite{Mayer2016}, for example, an encode-decoder architecture is adapted to directly predict a depth map based on a stereo image pair. To avoid learning the matching task from scratch, the authors further propose a variation of their basic architecture: First, the two images are processed individually in order to extract features, before a correlation layer is used to relate these features to find correspondences. The idea of mapping the single components of the classical dense stereo matching pipeline to a neural network architecture is further pursued in \cite{Kendall2017a}. After extracting features from the images individually, these features are used to build a cost volume which is optimised using 3D convolutions in an encoder decoder based structure. The final disparity map is obtained using a differentiable soft argmin operation. In \cite{Chang2018} on the other hand, the feature extraction step is extended by using spatial pyramid pooling. This approach allows to exploit global context by combining context from multiple scales resulting in more meaningful features.

\subsection{Uncertainty Estimation}

In the literature, a wide variety of different approaches for uncertainty estimation in the context of dense stereo matching, also referred to as confidence estimation, are presented. A majority of these methods operate based on a disparity map only \cite{Poggi2017, Tosi2018} or additionally take into account the RGB image used to compute the disparity map in the first place \cite{Fu2017}, considering the algorithm used to obtain the disparity map as block box. \cite{Veld2018} as well as \cite{Mehltretter2019} on the other hand, estimate the uncertainty using information contained in the 3D cost volume which is an intermediate representation present in most dense stereo matching algorithms. Benefiting from the additional information contained in such cost volumes compared to disparity maps, these methods allow a more accurate estimation of the uncertainty.

All the methods mentioned so far, interpret uncertainty estimation as a subsequent task to depth estimation. In consequence, the obtained uncertainty has no influence on the depth estimation process.
Learning to predict depth and uncertainty jointly, however, allows to use the obtained uncertainty as natural regularisation for the depth estimation, resulting in an improved accuracy.
Thus, in \cite{Shaked2017} and \cite{Kim2019} neural network architectures are proposed allowing to estimate depth and uncertainty jointly, based on a previously built cost volume. In addition, in \cite{Kendall2017} a neural network is trained end-to-end in the sense that it predicts depth and aleatoric uncertainty directly based on a stereo image pair.

In all of these methods, only aleatoric uncertainty is taken into account, while epistemic uncertainty is not considered. However, the epistemic part is crucial to approximate the true uncertainty of a depth prediction as good as possible. Furthermore, epistemic uncertainty delivers valuable information regarding a learned model being suitable or not to process a specific set of data, and indicates data samples which are outside of a learned distribution.
Therefore, an increasing number of approaches is presented recently, based on Bayesian models, which offer a mathematical framework to reason about model uncertainty and allow to estimate both, aleatoric and epistemic uncertainty \cite{Blundell2015, Gal2016, Wen2018}.
The ability of these methods to provide a good measure for uncertainty has already been demonstrated on various tasks such as monocular depth prediction and semantic segmentation \cite{Kendall2017b} as well as classification and regressing optical flow \cite{Gast2018}. However, the idea of estimating epistemic uncertainty via probabilistic deep learning is not yet applied to the task of dense stereo matching.
\section{Methodology}\label{sec:methodology}

In this section, the previously discussed idea of modelling both, aleatoric as well as epistemic uncertainty, is applied to the task of dense stereo matching. For the purpose of joint depth and uncertainty estimation from epipolar rectified stereo image pairs, we adapt a common neural network architecture, which is briefly outlined in Section \ref{sec:architecture}. While aleatoric uncertainty is directly predicted by the network, we infer epistemic uncertainty by transforming the network architecture into a probabilistic representation (c.f. Sec.\,\ref{sec:uncertainty_est}).

\subsection{Basic Architecture}\label{sec:architecture}

\begin{figure*}
    \centering
    \includegraphics[width=\linewidth]{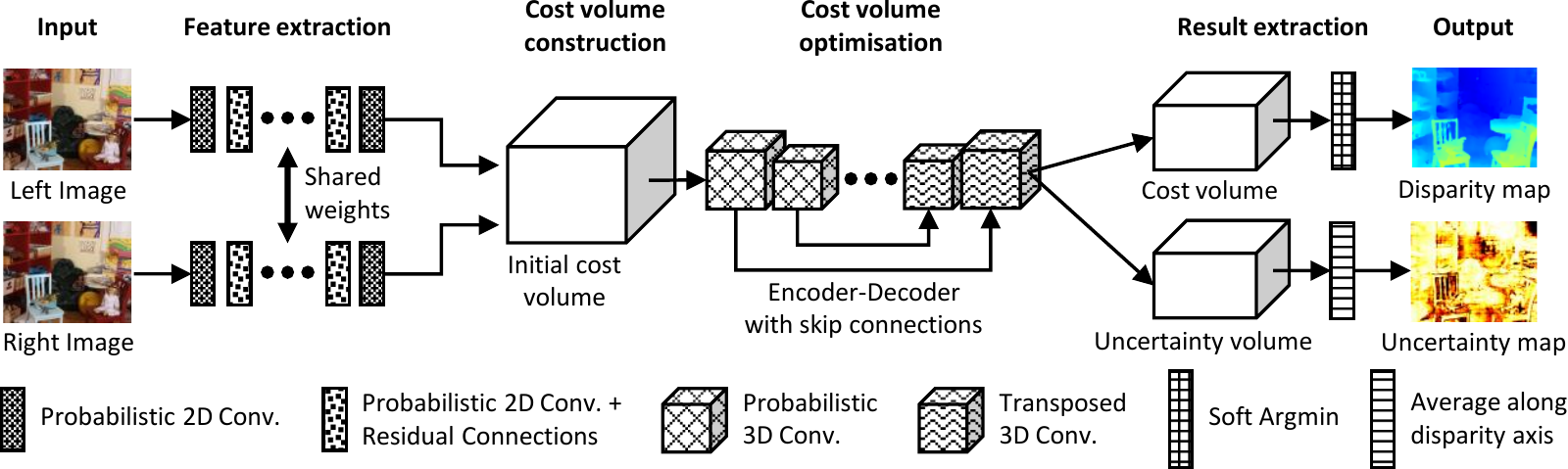}
    \caption{\textbf{Probabilistic GC-Net architecture.} Based on the GC-Net architecture \cite{Kendall2017a}, the 2D and 3D convolutional layers have been adapted to sample the corresponding parameters from a probability distribution instead of learning them directly.}
\label{fig:architecture}
\end{figure*}

As a basis of our subsequently proposed probabilistic neural network, the GC-Net architecture presented in \cite{Kendall2017a} is utilised, which consists of four major processing steps:

First, features are extracted from both images of a pair separately, using two branches of 2D convolutional layers arranged in residual blocks. In order to force the network to extract similar features from both images, the weights from both branches are shared. 
In the second step, a cost volume is built by concatenating features from the left image with features from the right image along the corresponding horizontal epipolar line and for all disparity levels within a specified range.
The resulting 4D volume is further processed using 3D convolutional and transposed convolutional layers arranged in an encoder-decoder structure with skip connections. This allows the network to optimise the cost volume on different scales and with a wide field of view. In the last step, a disparity map is extracted from the optimised cost volume using a differentiable soft argmin layer.

Following the modifications proposed in \cite{Kendall2017}, the network is enabled to also predict aleatoric uncertainty. For this purpose, the last 3D transposed convolutional layer is modified to predict two 3D volumes instead of one. The first volume contains the cost information, the second one the corresponding uncertainties. The aleatoric uncertainty map is obtained by averaging the uncertainty values along the disparity axis.

The GC-Net architecture demonstrates good accuracy for the tasks of depth and aleatoric uncertainty estimation, while having a relatively low number of parameters ($\sim$2.8 Mio.), mainly justified by the absence of fully-connected layers. The low number of parameters is especially important in the context of our goal to transform the architecture into a probabilistic representation, since this procedure may multiply the number of parameters, depending on the distribution types used.

\subsection{Epistemic Uncertainty Estimation}\label{sec:uncertainty_est}
Since the reviewed GC-Net architecture only allows to estimate aleatoric uncertainty, in the present work we transfer this architecture from a deterministic to a probabilistic representation (c.f. Fig.\,\ref{fig:architecture}). Using such a probabilistic representation, we enable the network to additionally estimate epistemic uncertainty.
In more detail, we realise a Bayesian approach, where the weights $W$ of 2D and 3D convolutional layers are sampled from a probability distribution $q_{\Theta}$ which is optimised during training, instead of learning the weights directly. Consequently, for every prediction a different set of weights is sampled, resulting in varying disparity maps for the same image pair. Since the differences between these disparity maps result from the model's uncertainty to assign disparities to certain pixels, these variations are used to approximate the epistemic uncertainty.

For this purpose, the network predicts a disparity $\hat{d}$ and a variance $\hat{\sigma}^2$ for every pixel $i$ of the left input image. For every prediction $t$ a set of weights $W$ is sampled from the distribution $q_{\Theta}$ with learned parameters $\Theta$, resulting in a Monte Carlo sampling approach.
To obtain the final disparity estimation $ \bar{d}$, the prediction is repeated $T$ times and the results are averaged:
\begin{equation}\label{eq:disparity}
    \bar{d}_i=\frac{1}{T}\sum_{t=1}^{T}\hat{d}_{i,t}\,.
\end{equation}
The final variance which corresponds to a disparity $d_i$ is approximated according to \cite{Kendall2017b}, by simply adding epistemic and aleatoric uncertainty:
\begin{equation}\label{eq:variance}
    \text{Var}(d_i)\approx\frac{1}{T}\sum_{t=1}^{T}(\hat{d}_{i,t}-\bar{d}_i)^{2}+\frac{1}{T}\sum_{t=1}^{T}\hat{\sigma}_{i,t}^2\,,
\end{equation}
where the first part, the variance of the predicted disparities, represents epistemic and the second part, the average of the predicted variances $\hat{\sigma}^2$, represents aleatoric uncertainty.

In this work, we realise $q_{\Theta}$ via the combination of normal distributions. Consequently, for every weight that is sampled from such a normal distribution instead of being learned directly, a mean and a variance need to be learned, which doubles the number of parameters. To keep this number as low as possible, only 2D and 3D convolutions are replaced, 3D transposed convolutions, on the other hand, are retained deterministically. This procedure allows the feature extraction (2D convolutions) and multi-scale feature matching (3D convolutions) steps to produce varying results, while keeping the feature map upscaling operations (3D transposed convolutions), carried out in the decoder part of the cost volume optimisation step, constant.

\subsection{Loss Function}

The proposed probabilistic neural network is trained end-to-end in a supervised manner, assuming ground truth disparity to be known. For this purpose, a two-part loss function is used:
\begin{equation}
    \mathcal{L}(d,\hat{d},\Theta) = \mathcal{L}_{Reg}(d,\hat{d}) + \mathcal{L}_{Prob}(\Theta)\,,
\end{equation}
where $\mathcal{L}_{Reg}$ describes the regression task itself, while $\mathcal{L}_{Prob}$ regularises the probabilistic components of our neural network architecture. Moreover, $d$ is the reference disparity, $\hat{d}$ the estimated disparity and $\Theta$ are the parameters of the network. 
The regression part of the loss function is taken from \cite{Kendall2017} and is defined as:
\begin{equation}
    \mathcal{L}_{Reg}(d,\hat{d}) = \frac{1}{N}\sum_{i=1}^{N}\frac{1}{2}\exp(-s_i)||d_i-\hat{d}_i||+\frac{1}{2}s_i\,,
\end{equation}
where the log variance $s_i=\log\hat{\sigma}_i^2$ is predicted instead of regressing the variance $\hat{\sigma}_{i}^2$ directly. This procedure is numerically more stable and prevents the loss from being divided by zero. Furthermore, the predicted aleatoric uncertainty is used as a weight for the difference between estimated and reference disparity in order to balance the influence of difficult and simple samples, respectively. Finally, the loss is averaged over all $N$ pixels processed in a single forward pass.

The second part of our loss function uses the Kullback-Leibler divergence $KL$ \cite{Kullback1951} as regularisation term:
\begin{equation}
    \mathcal{L}_{Prob}(\Theta)=KL(q_{\Theta}||p)\,,
\end{equation}
where $p$ is the real but intractable probability distribution of our probabilistic model and $q_{\Theta}$ is an approximation of the real distribution, which is described by a set of parameters $\Theta$. During training, $q_{\Theta}$ is optimised using variational inference \cite{Graves2011} to minimise the Kullback-Leibler divergence up to a constant, also referred to as negative Evidence Lower Bound.
\section{Experimental Results}\label{sec:eval}

Within this section, the previously introduced probabilistic deep learning approach is analysed and evaluated. For this purpose, the results of three different datasets are assessed, which are introduced in Section \ref{sec:datasets}. The estimated depth and uncertainty values are evaluated and compared against the results of the original network (GC-Net) \cite{Kendall2017a} and a modified version which additionally estimates aleatoric uncertainty (GC-Net-A) \cite{Kendall2017}. To allow a direct comparison, both variants are trained following the same strategy used for training the proposed probabilistic neural network (c.f. Sec.\,\ref{sec:training}).

\begin{figure}
    \centering
    \includegraphics[width=0.98\linewidth]{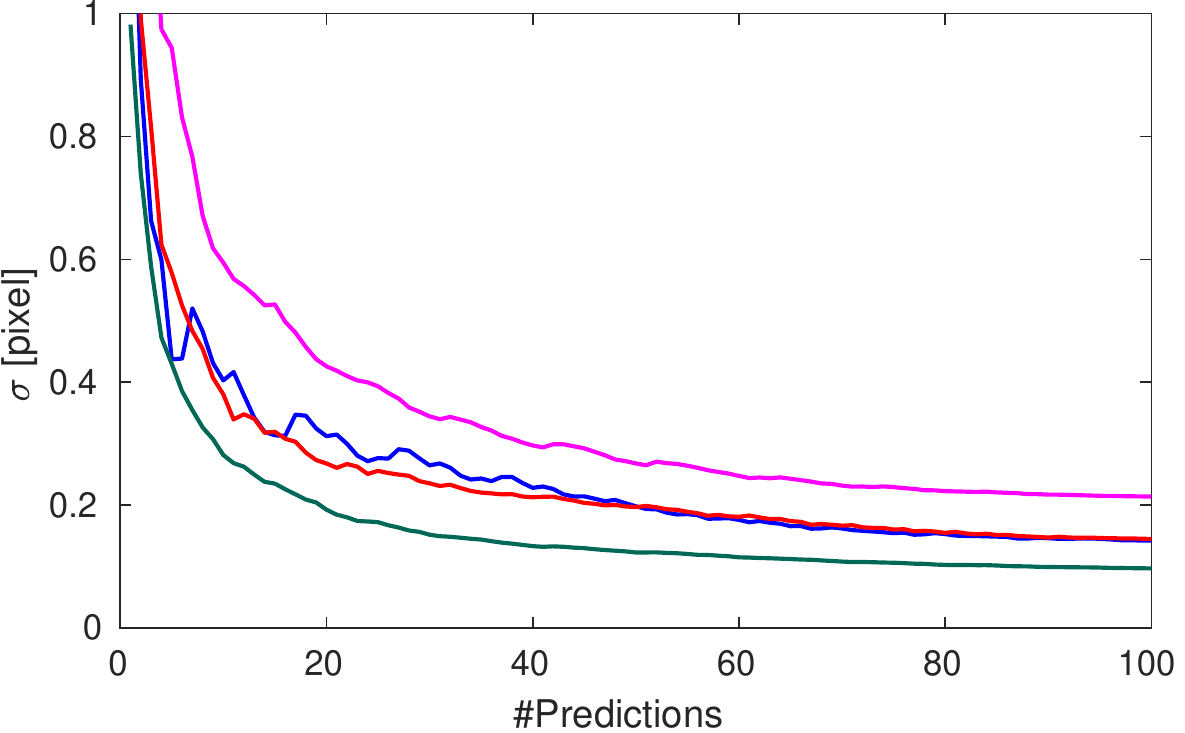}
    \caption{\textbf{Necessary number of predictions during test time.} With increasing number of predictions $T$ used for estimating a disparity map $\bar{d}_i$ (c.f. Eq.\,\ref{eq:disparity}), the average pixel-wise standard deviation of $i$ such estimations decreases and converges against the epistemic uncertainty. Here this effect is shown for $i=10$ on four different image pairs.}
\label{fig:prediction-ana}
\end{figure}

For all experiments, the number of predictions performed to estimate a disparity and an uncertainty map using our probabilistic neural network is set to $T=50$ (c.f. Eq.\,\ref{eq:disparity} and Eq.\,\ref{eq:variance}). As can be seen in Figure \ref{fig:prediction-ana}, an increasing number of predictions improves the stability of the estimated disparity maps, in the sense that the standard deviation of multiple estimations for the same stereo image pair decreases. At the same time, however, an increased number of predictions also leads to a higher computational effort and thus to a longer processing time. In consequence, performing 50 predictions seems to be a reasonable trade-off, since only minor improvements can be achieved by running additional predictions.

\subsection{Datasets}\label{sec:datasets}

For the experiments carried out in the context of this work, three datasets are used, which all consist of stereo image pairs with known reference depth: Sceneflow FlyingThings3D \cite{Mayer2016}, KITTI 2015 \cite{Menze2015} and Middlebury v3 \cite{Scharstein2014}.
The Sceneflow dataset contains more than 25k synthetic stereo image pairs showing a high diversity of different scenes. For all images, sub-pixel accurate dense reference disparity maps are available. 
On the other hand, the KITTI dataset contains real image pairs, which were captured using vehicle mounted stereo camera set-ups and provides LIDAR based ground truth disparity maps with disparities for 30\,\% of the pixels. Containing various street scenes from urban as well as rural environments, this dataset still pose a challenge to dense stereo matching algorithms.
Finally, the Middlebury dataset contains 15 image pairs showing various indoor scenes captured with a static stereo set-up and providing dense ground truth disparity maps based on structured light. Due to hardware limitations, the images of the Middlebury dataset are processed at one quarter of the original resolution, within this evaluation.

\subsection{Training Procedure}\label{sec:training}
To train the proposed probabilistic neural network, the stereo image pairs contained in the Sceneflow dataset are used. From these image pairs, random extracts of size $256\times128$ are cropped and fed to the network during training. Using a batch size of $1$ the network is trained for $12$ epochs, in the sense that one extract from every image pair is seen per epoch.
While the parameters of the probabilistic 2D and 3D convolutional layers are updated via variational inference using Flipout \cite{Wen2018}, for the 3D transposed convolutional layers Glorot initialisation \cite{Glorot2010} is used. 
Finally, RMSProb with a learning rate of $1\times10^{-3}$ is employed for optimisation. 

\subsection{Depth Evaluation}\label{sec:depth-eval}

\begin{table*}
	\centering
	\footnotesize
	\begin{tabular}{l|ccccc|cc}
		\hline
		\  & \multicolumn{5}{c|}{Error} & \multicolumn{2}{c}{Uncertainty}\\
        \ & $>1\,px [\%]$ & $>3\,px [\%]$ & $>5\,px [\%]$ & MAE $[px]$ & RMSE $[px]$ & Aleatoric $[px]$ & Epistemic $[px]$ \\\hline\hline
        \multicolumn{8}{c}{\textbf{Sceneflow} \cite{Mayer2016}}\\\hline
        \ GC-Net & 13.9 & 8.0 & 6.5 & 10.15 & 14.99 & - & - \\
        \ GC-Net w. Aleatoric Uncertainty & 15.8 & 10.1 & 8.5 & 9.98 & 18.32 & 2.95 & - \\
        \ Probabilistic GC-Net (ours) & 15.2 & 9.3 & 7.6 & 9.47 & 14.88 & 2.10 & 2.65 \\\hline
        \multicolumn{8}{c}{\textbf{KITTI 2015} \cite{Menze2015}}\\\hline
        \ GC-Net & 57.9 & 25.9 & 16.4 & 4.34 & 11.62 & - & - \\
        \ GC-Net w. Aleatoric Uncertainty & 86.7 & 69.9 & 62.1 & 47.27 & 75.43 & 6.57 & - \\
        \ Probabilistic GC-Net (ours) & 83.5 & 62.4 & 52.7 & 22.29 & 39.72 & 4.61 & 11.56 \\\hline
        \multicolumn{8}{c}{\textbf{Middlebury v3} \cite{Scharstein2014}}\\\hline
        \ GC-Net & 30.4 & 15.5 & 11.3 & 4.10 & 10.33 & - & - \\
        \ GC-Net w. Aleatoric Uncertainty & 40.5 & 25.7 & 21.2 & 9.94 & 25.36 & 3.71 & - \\
        \ Probabilistic GC-Net (ours) & 34.5 & 19.8 & 15.0 & 4.38 & 10.48 & 3.08 & 3.85 \\\hline
		\hline
    \end{tabular}
    \caption{\textbf{Quantitative results on the three evaluated datasets.} From the Sceneflow and the KITTI  dataset, 100 random image pairs and from the Middlebury dataset, all 15 training pairs are used for evaluation. The results are compared against the original GC-Net architecture \cite{Kendall2017a} and a modified variant which allows to estimate aleatoric uncertainty \cite{Kendall2017}.}
    \label{table:results}
\end{table*}

In the first part of the evaluation, the accuracy of the disparity maps estimated by the proposed probabilistic neural network is assessed and compared to the results of the original GC-Net \cite{Kendall2017a} and a variant which additionally estimates aleatoric uncertainty \cite{Kendall2017}. Analysing the results presented in Table \ref{table:results}, it can be seen that our probabilistic approach estimates disparity maps for the Sceneflow dataset with an accuracy comparable to the one of the original approach. While the percentages of "bad" pixels (deviation of the estimated from the reference disparity is larger than a specified threshold) is slightly higher for the probabilistic approach, the mean average error (MAE) as well as the root mean square error (RMSE) are a bit lower. A similar behaviour can be observed on the Middlebury dataset.

\begin{figure*}[thb]
\captionsetup[subfigure]{labelformat=empty}
\centering
\subfloat
{
    \includegraphics[width=0.24\linewidth]{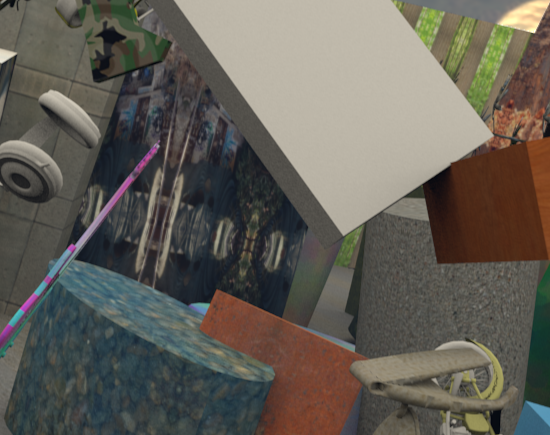}
}\hfill
\subfloat
{
    \includegraphics[width=0.24\linewidth]{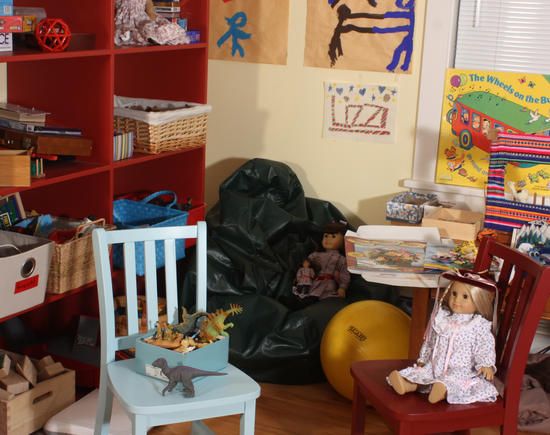}
}\hfill
\subfloat
{
    \includegraphics[width=0.48\linewidth]{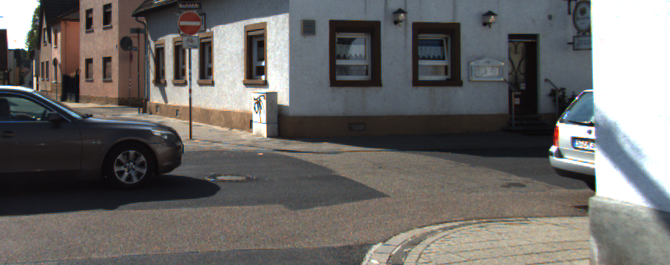}
}\newline
\subfloat
{
    \includegraphics[width=0.24\linewidth]{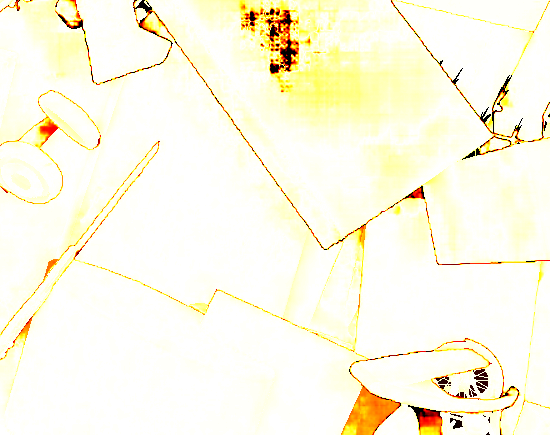}
}\hfill
\subfloat
{
    \includegraphics[width=0.24\linewidth]{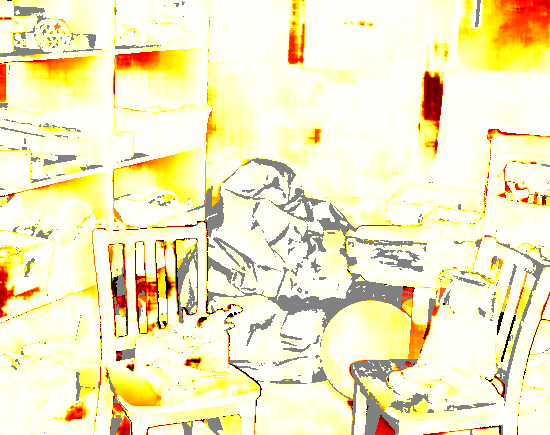}
}\hfill
\subfloat
{
    \includegraphics[width=0.48\linewidth]{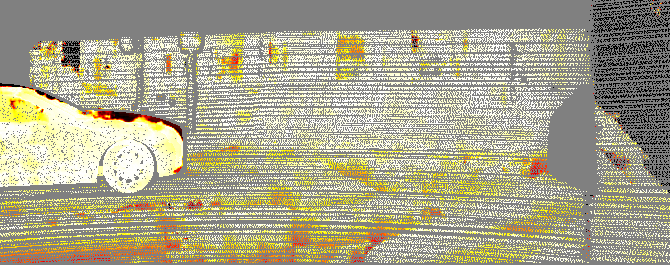}
}\newline
\subfloat
{
    \includegraphics[width=0.24\linewidth]{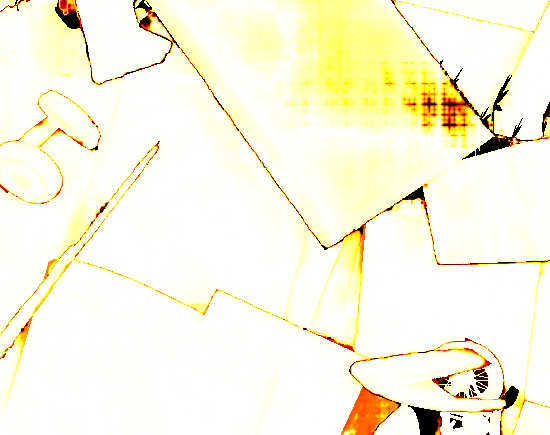}
}\hfill
\subfloat
{
    \includegraphics[width=0.24\linewidth]{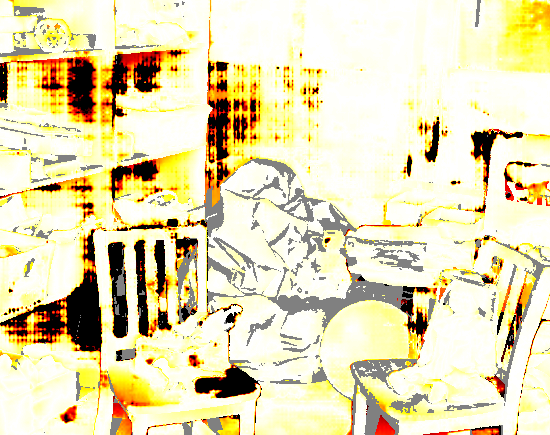}
}\hfill
\subfloat
{
    \includegraphics[width=0.48\linewidth]{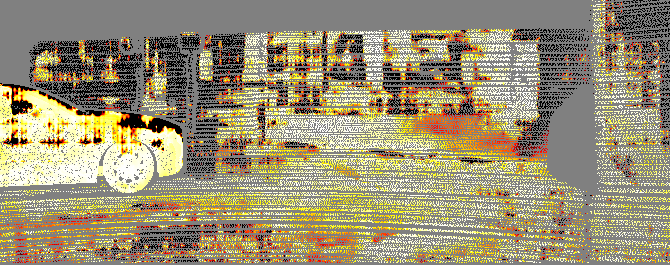}
}\newline
\subfloat
{
    \includegraphics[width=0.24\linewidth]{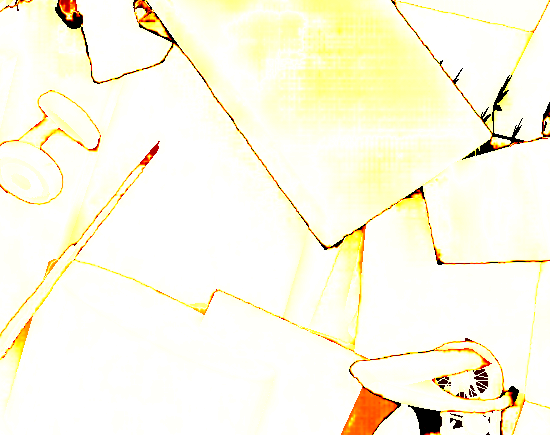}
}\hfill
\subfloat
{
    \includegraphics[width=0.24\linewidth]{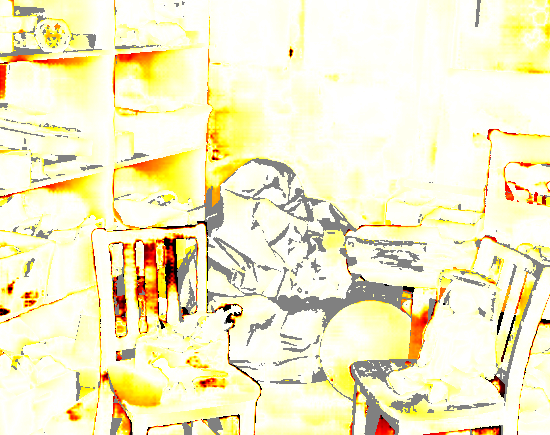}
}\hfill
\subfloat
{
    \includegraphics[width=0.48\linewidth]{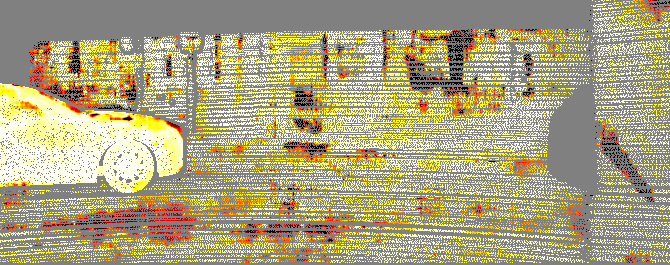}
}
\caption{\textbf{Qualitative depth evaluation.} From left to right, one example from the Sceneflow FlyingThings3D, the Middlebury v3 and the KITTI 2015 dataset is shown. 
From top to bottom, the left reference image 
and the error maps for the original GC-Net \cite{Kendall2017a}, GC-Net with aleatoric uncertainty \cite{Kendall2017} and the probabilistic GC-Net, proposed in this work, are shown. 
The error maps encode a high error in black and a small one in white, while pixels without ground truth are displayed in grey. In general, it can seen that the original GC-Net architecture outperforms our probabilistic variant in areas with fine details, while the probabilistic architecture demonstrates superior performance in low textured areas.}
\label{fig:results-disp}
\end{figure*}

The qualitative results shown in Figure \ref{fig:results-disp} illustrate that the original GC-Net is superior in estimating the correct disparity for fine structures, while our probabilistic approach tends to over-smooth such details. For large low-textured areas, on the other hand, our method is able to estimate the correct disparity for a majority of the pixels, while the predictions of the original network often contain artefacts, visible in all three examples.
Since the described effect of over-smoothing fine structures is also present in the disparity maps of GC-Net-A, the reason may probably be found in the definition of the loss function. Weighting the disparity difference with the predicted uncertainty minimises the influence of challenging samples on the loss function, such as pixels close to depth discontinuities. Consequently, the learned model is less accurate for predicting disparities in such regions. To overcome this limitation, an adjustment of the loss function may be beneficial, e.g. by introducing gradient information, as suggested in \cite{Kang2019a}.

On the KITTI dataset, the percentage of bad pixels as well as the MAE and the RMSE increase significantly for all three methods compared to the Sceneflow dataset. This shows that the trained models are not able to generalise well from the training to the test domain, indicating a significant difference in the distributions of the two datasets. However, the original GC-Net suffers much less from this effect than the two others. It is noticeable that our probabilistic approach as well as GC-Net-A have particular difficulties to estimate the correct disparity for small objects in greater distances and for vegetation, both being underrepresented in the training samples or not contained at all. 

Since the original GC-Net is less sensitive to the domain gap between the different analysed datasets, the modifications made on the architecture and the loss function to not only obtain disparity but also aleatoric uncertainty information, seem to result in an over-fitting effect on the training domain. While our probabilistic modifications mitigate this effect, it is still visible in the results. In consequence, further investigations have to be carried out in the future to analyse this behaviour in more detail.

\begin{figure}[t]
    \centering
    \subfloat[GC-Net with aleatoric uncertainty]
    {
        \includegraphics[width=0.98\linewidth]{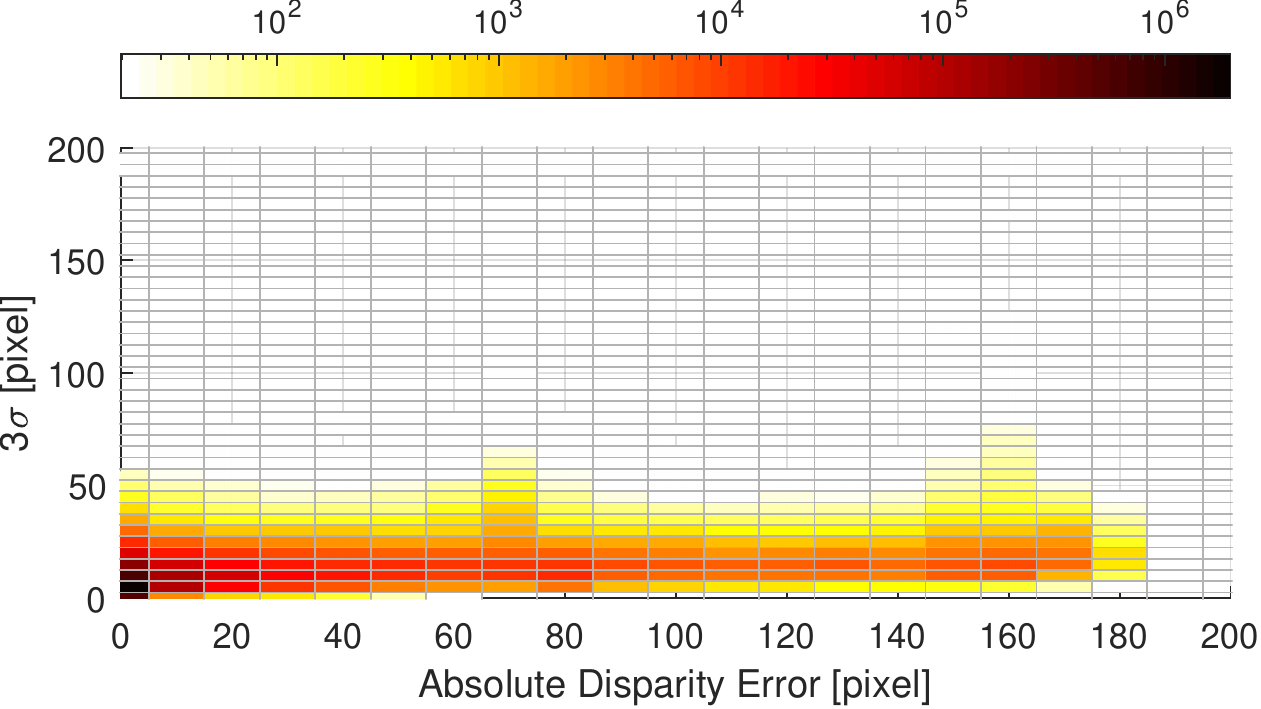}
    }\newline
    \subfloat[Ours]
    {
        \includegraphics[width=0.98\linewidth]{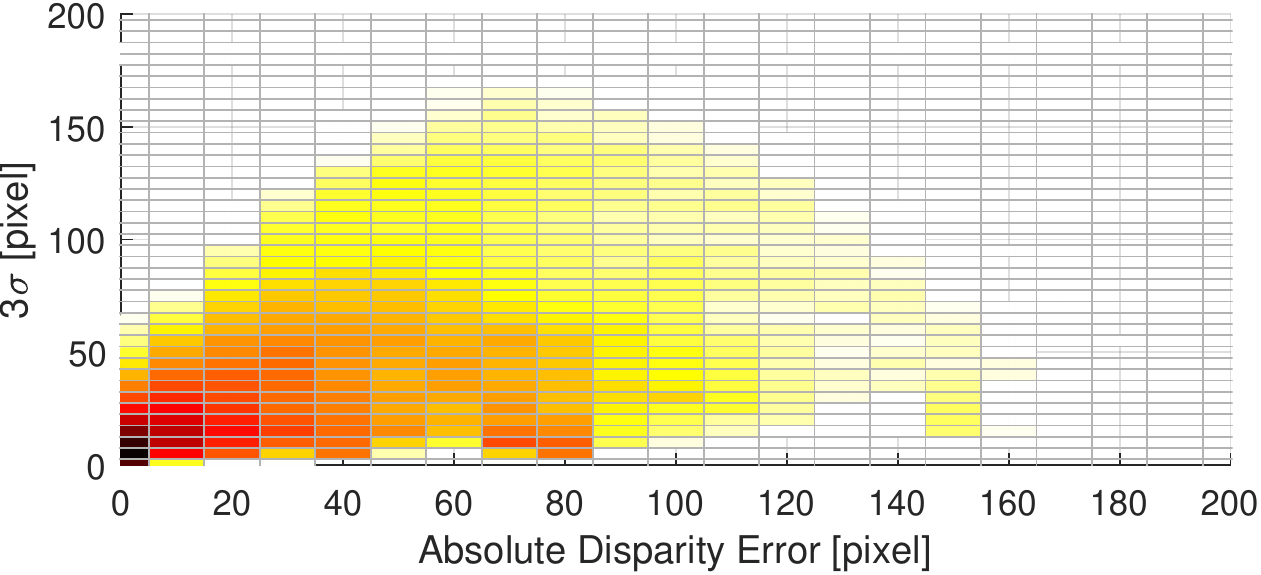}
    }
    \caption{\textbf{Absolute error uncertainty relation on all images of the Middlebury dataset.} The logarithmic colour scale encodes the number of pixels having the respective error and estimated standard deviation $\sigma$. In case of optimal uncertainty estimation, all pixels would be located close to the diagonal, showing an increasing uncertainty with increasing error.}
    \label{fig:abs-error-unc-rel}
\end{figure}

\begin{figure}[t]
    \centering
    \includegraphics[width=0.98\linewidth]{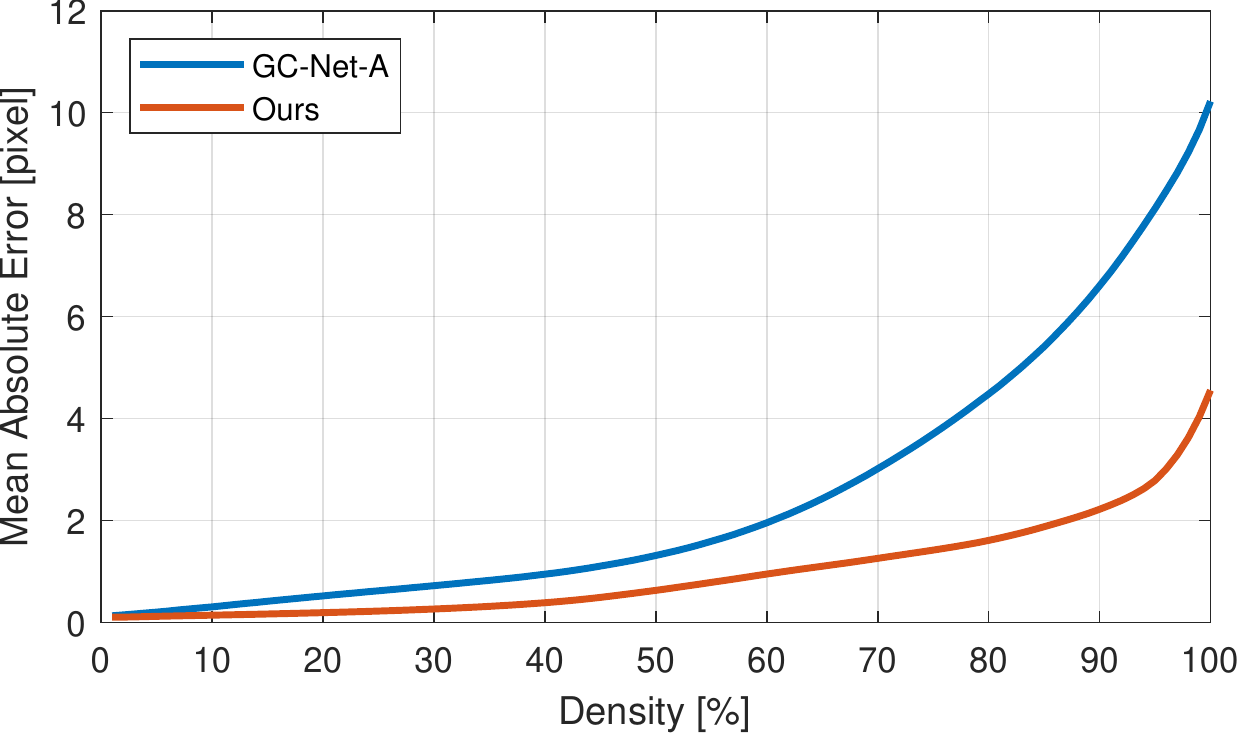}
    \caption{\textbf{Relative error uncertainty relation on all images of the Middlebury dataset.} A curve represents the mean absolute error as a function of the percentage of pixels sampled from a disparity map in order of increasing uncertainty. In case of an optimal uncertainty estimation, this procedure would be equal to sample pixels in the order of increasing error, resulting in a minimal area under the curve.}
    \label{fig:rel-error-unc-rel}
\end{figure}

\subsection{Uncertainty Evaluation}

\begin{figure*}[thb]
\captionsetup[subfigure]{labelformat=empty}
\centering
\subfloat
{
    \includegraphics[width=0.24\linewidth]{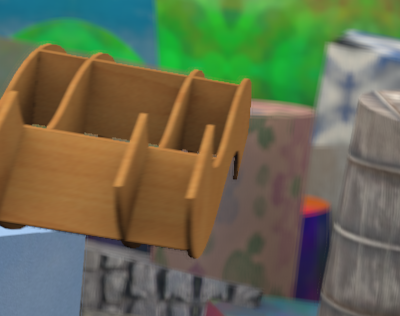}
}\hfill
\subfloat
{
    \includegraphics[width=0.24\linewidth]{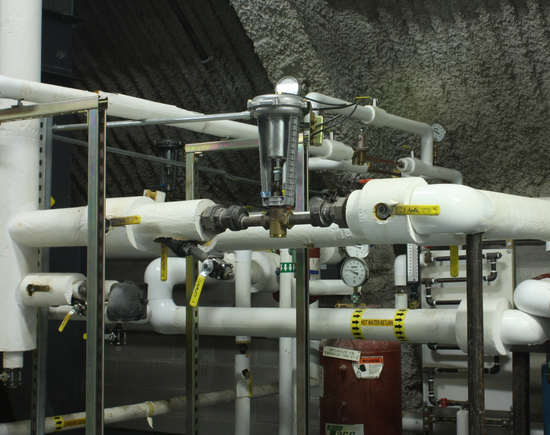}
}\hfill
\subfloat
{
    \includegraphics[width=0.48\linewidth]{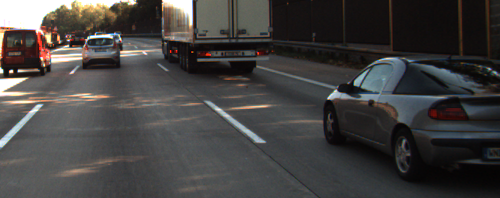}
}\newline
\subfloat
{
    \includegraphics[width=0.24\linewidth]{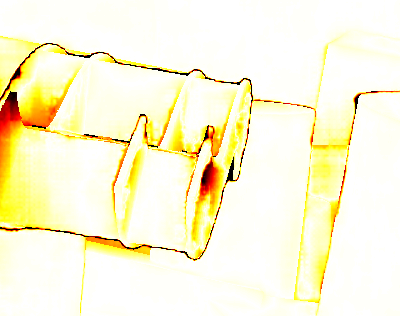}
}\hfill
\subfloat
{
    \includegraphics[width=0.24\linewidth]{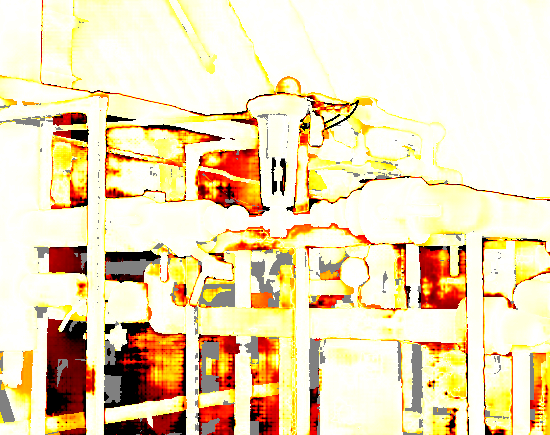}
}\hfill
\subfloat
{
    \includegraphics[width=0.48\linewidth]{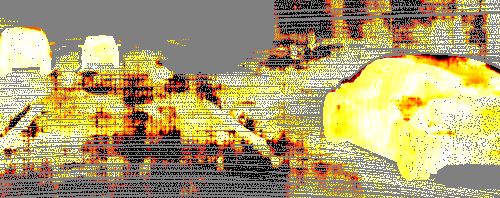}
}\newline
\subfloat
{
    \includegraphics[width=0.24\linewidth]{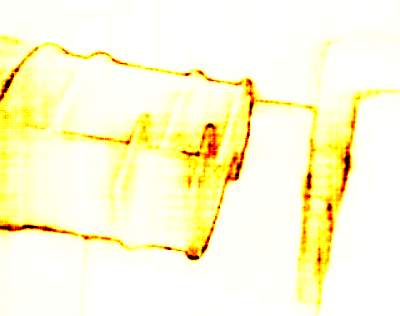}
}\hfill
\subfloat
{
    \includegraphics[width=0.24\linewidth]{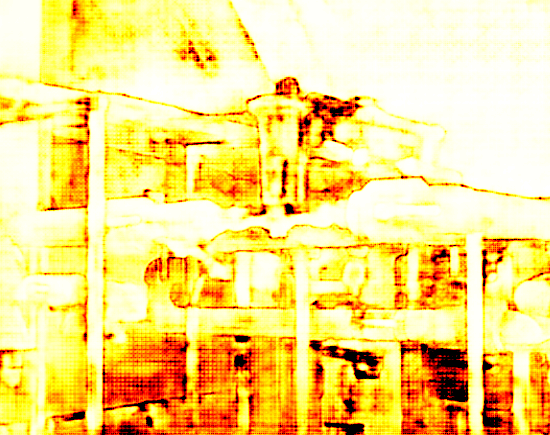}
}\hfill
\subfloat
{
    \includegraphics[width=0.48\linewidth]{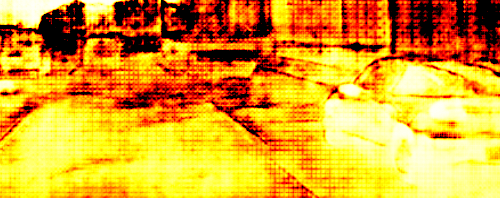}
}\newline
\subfloat
{
    \includegraphics[width=0.24\linewidth]{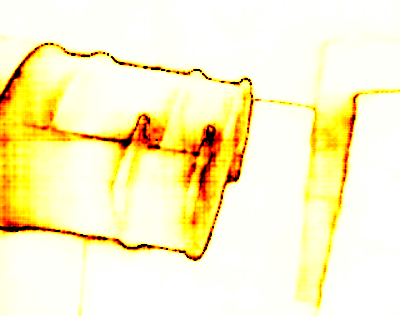}
}\hfill
\subfloat
{
    \includegraphics[width=0.24\linewidth]{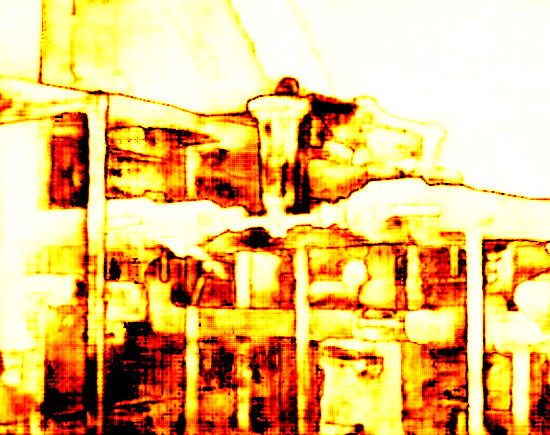}
}\hfill
\subfloat
{
    \includegraphics[width=0.48\linewidth]{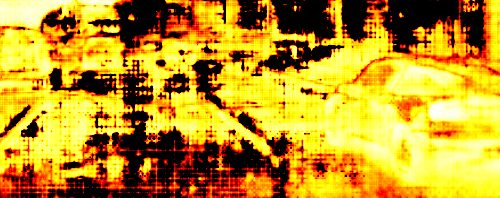}
}
\caption{\textbf{Qualitative uncertainty evaluation.} From left to right, one example from the Sceneflow FlyingThings3D, the Middlebury v3 and the KITTI 2015 dataset is shown. From top to bottom, the left reference image, the error map of the proposed probabilistic neural network and the uncertainty maps for aleatoric uncertainty alone and for aleatoric and epistemic uncertainty together are shown.
The error map and the uncertainty maps encode a high value in black and a small one in white, while pixels without ground truth disparity are displayed in grey. All three examples show that the additional modelling of epistemic uncertainty allows to better approximate the real error distribution. The example of the KITTI dataset in particular demonstrates that epistemic uncertainty is crucial to identify samples for which the learned model is unable to make a correct prediction due to a domain gap.}
\label{fig:results-unc}
\end{figure*}

In the last part of the evaluation, the estimated uncertainty of the proposed probabilistic approach is assessed and compared against the results of GC-Net-A \cite{Kendall2017}. As shown in Table \ref{table:results}, our approach predicts a lower aleatoric but a higher combined uncertainty (c.f. Eq.\,\ref{eq:variance}) for all three evaluated datasets. This indicates that if only aleatoric uncertainty is modelled, the model tries to compensate effects which originate from the model uncertainty. Modelling both, aleatoric and epistemic uncertainty, however, allows a better differentiation of the effects coming from data and model uncertainty.

Analysing the qualitative results shown in Figure \ref{fig:results-unc}, it can be seen that GC-Net-A often underestimates the real uncertainty. 
This behaviour is also visible in Figure \ref{fig:abs-error-unc-rel}, which shows the pixel-wise relation between the absolute disparity error and the estimated uncertainty. While an optimal uncertainty estimation procedure would result in points close to the diagonal, GC-Net-A often assigns a low uncertainty even to pixels with a high error, which results in a horizontal distribution. Our probabilistic approach, on the other hand, approximates the real error distribution much more accurately.

Estimating epistemic uncertainty has the additional benefit that it directly indicates how well the trained network is suited to predict disparity maps for images of a specific test domain. As can be seen in Table \ref{table:results}, the epistemic uncertainty is relatively small for the Sceneflow dataset, which was also used for training the network. On the Middlebury dataset the epistemic uncertainty is a bit higher, indicating that the contained image pairs show scenes which slightly differ from the training set. For the KITTI dataset, on the other hand, the epistemic uncertainty is significantly higher, indicating that the characteristics of this dataset differs greatly from those of the training samples. While the Sceneflow dataset contains images showing synthetic scenes, the Middlebury images show indoor environments with controlled external influences and the KITTI dataset shows road scenes with varying external influences.
Consequently, the actual differences between the datasets agree with the assumptions implied by the estimated epistemic uncertainties.

Analysing the qualitative results shown in Figure \ref{fig:results-unc} in more detail, it is noticeable that our approach estimates high uncertainty in particular close to object borders, and therefore at depth discontinuities, and for fine structures. This is in accordance with the observations that our probabilistic approach tends to over-smooth estimated disparity maps, as described in Section \ref{sec:depth-eval}. This indicates that the majority of erroneous disparity predictions concur with high uncertainty estimations and can therefore be detected as errors by our method.
This assumption is supported by the results shown in Figure \ref{fig:rel-error-unc-rel}, which visualises the mean absolute error as a function of the percentage of pixels sampled from a disparity map in order of increasing uncertainty. 
While the curve that corresponds to our method increases slowly until about $95\%$ density, it has a much higher slope afterwards, illustrating that the majority of pixels with a high error are sampled in this segment, which means that these pixels have also been assigned a high uncertainty.
Especially for the KITTI example this is noteworthy: While the accuracy of the proposed probabilistic method is significantly worse compared to the results of the original GC-Net, most of the erroneous disparity predictions can be identified based on the combined uncertainty provided by our method. This is not the case if only aleatoric uncertainty is estimated, since many incorrect depth predictions are missed, especially on the street in the foreground and for the vegetation in the upper right corner.
\section{Conclusion}
Inspired by the recent progress in the field of probabilistic deep learning, within the present work, we propose an end-to-end trained probabilistic neural network to jointly estimate depth and aleatoric as well as epistemic uncertainty from epipolar rectified stereo image pairs.
Within an extensive evaluation using three well established and commonly used datasets we assess the quality of the estimated depth and uncertainty information. It is shown that in most of the examples evaluated, the probabilistic modifications only have a minor influence on the overall accuracy of the estimated disparity maps. However, the tendency to over-fit on the training domain is increased and fine structures are more often over-smoothed. These limitations are subject to further investigations and may be mitigated by an adjustment of the loss function, e.g. via the introduction of gradient information. Furthermore, the use of other distributions than the normal distribution for sampling the network weights might be beneficial and will be investigated.

On the other hand, the estimated disparity maps show an improved accuracy in low textured areas, visible by the reduced amount of noise and artefacts.
Furthermore, the estimated uncertainty information shows superior quality compared to a variant, which predicts aleatoric uncertainty only, and allows to identify a large majority of erroneous disparity estimations. At the same time, the importance of also modelling epistemic uncertainty is demonstrated: It not only allows a better approximation of the real error distribution, but provides a direct indicator how well a learned model for depth estimation is suited to predict disparity maps for images from a specific test domain.

\section*{Acknowledgements}
This work was supported by the German Research Foundation (DFG) as a part of the Research Training Group i.c.sens [GRK2159], the MOBILISE initiative of the Leibniz University Hannover and TU Braunschweig and by the NVIDIA Corporation with the donation of the Titan V GPU used for this research.

{
	\begin{spacing}{1.17}
		\normalsize
		\bibliography{LiteraturBibliothek}
	\end{spacing}
}

\end{document}